\documentclass[conference]{IEEEtran}
\IEEEoverridecommandlockouts

\usepackage{cite}
\usepackage{amsmath,amssymb,amsfonts}
\usepackage{algorithmic}
\usepackage{graphicx}
\usepackage{textcomp}
\usepackage{xcolor}
\def\BibTeX{{\rm B\kern-.05em{\sc i\kern-.025em b}\kern-.08em
    T\kern-.1667em\lower.7ex\hbox{E}\kern-.125emX}}

\usepackage{setspace}
\usepackage{multirow}
\usepackage{makecell}
\usepackage{booktabs}
\usepackage{color}
\usepackage{url}
\usepackage{siunitx}
\usepackage{booktabs}
\usepackage{amsmath} 
\usepackage{arydshln}

\makeatletter
\newcommand{\linebreakand}{%
  \end{@IEEEauthorhalign}
  \hfill\mbox{}\par
  \mbox{}\hfill\begin{@IEEEauthorhalign}
}
\makeatother

\begin{document}

\title{Explore the Hallucination on Low-level \\Perception for MLLMs\\
\thanks{$^{*}$Corresponding authors.}
}

\author{
\IEEEauthorblockN{Yinan Sun}
\IEEEauthorblockA{\textit{Shanghai Jiao Tong University} \\
}
\and
\IEEEauthorblockN{Zicheng Zhang}
\IEEEauthorblockA{\textit{Shanghai Jiao Tong University} \\
}
\and
\IEEEauthorblockN{Haoning Wu}
\IEEEauthorblockA{\textit{S-Lab, Nanyang Technological University} \\
}

\linebreakand 
\IEEEauthorblockN{Xiaohong Liu}
\IEEEauthorblockA{\textit{Shanghai Jiao Tong University} \\
}
\and
\IEEEauthorblockN{Weisi Lin}
\IEEEauthorblockA{\textit{Nanyang Technological University} \\
}

\linebreakand 
\IEEEauthorblockN{Guangtao Zhai$^{*}$}
\IEEEauthorblockA{\textit{Shanghai Jiao Tong University} \\
Shanghai, China \\
zhaiguangtao@sjtu.edu.cn}
\and
\IEEEauthorblockN{Xiongkuo Min$^{*}$}
\IEEEauthorblockA{\textit{Shanghai Jiao Tong University} \\
Shanghai, China \\ 
minxiongkuo@sjtu.edu.cn}
\vspace{-2em}
}

\maketitle
\vspace{-2em}
\begin{abstract}
The rapid development of Multi-modality Large Language Models (MLLMs) has significantly influenced various aspects of industry and daily life, showcasing impressive capabilities in visual perception and understanding. However, these models also exhibit hallucinations, which limit their reliability as AI systems, especially in tasks involving low-level visual perception and understanding. We believe that hallucinations stem from a lack of explicit self-awareness in these models, which directly impacts their overall performance. In this paper, we aim to define and evaluate the self-awareness of MLLMs in low-level visual perception and understanding tasks. To this end, we present QL-Bench, a benchmark settings to simulate human responses to low-level vision, investigating self-awareness in low-level visual perception through visual question answering related to low-level attributes such as clarity and lighting. Specifically, we construct the LLSAVisionQA dataset, comprising 2,990 single images and 1,999 image pairs, each accompanied by an open-ended question about its low-level features. Through the evaluation of 15 MLLMs, we demonstrate that while some models exhibit robust low-level visual capabilities, their self-awareness remains relatively underdeveloped. Notably, for the same model, simpler questions are often answered more accurately than complex ones. However, self-awareness appears to improve when addressing more challenging questions. We hope that our benchmark will motivate further research, particularly focused on enhancing the self-awareness of MLLMs in tasks involving low-level visual perception and understanding.
\end{abstract}


\begin{figure}[!t]
\centerline{\includegraphics[width=3.5in]{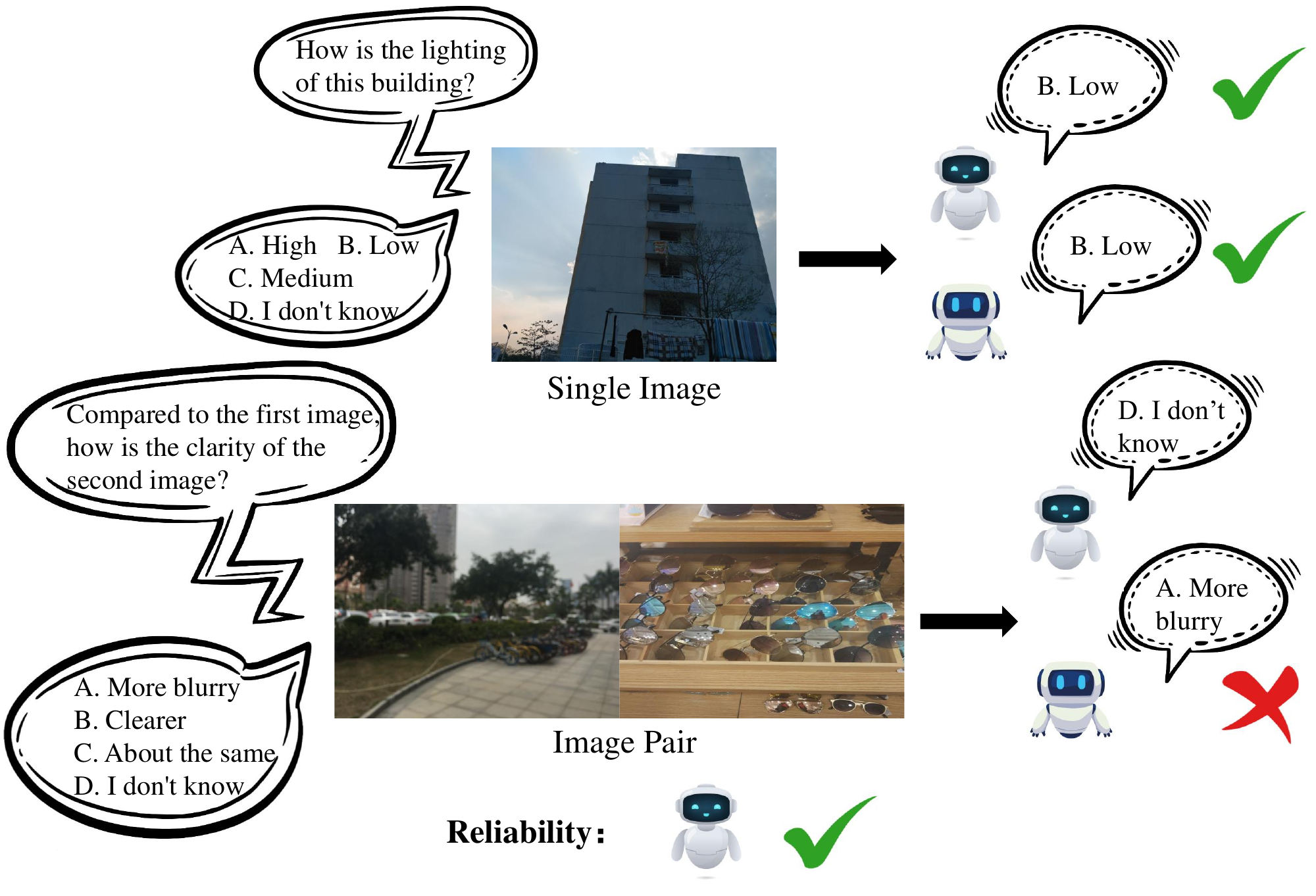}}
\caption{Self-awareness of MLLMs in low-level vision tasks. A trustworthy MLLM can accurately identify what it knows and what it does not know.}
\label{example}
\vspace{-1em}
\end{figure}

\section{Introduction}
In recent years, the rapid development of Large Language Models (LLMs), such as close-source models like GPT and Bard, open-source models like LLaMA\cite{touvron2023llama} and MPT\cite{mosaicml2023introducing}, has significantly influenced the direction of AI development. To better align with practical application, researchers have integrated multi-modality capabilities into LLMs, giving rise to Multi-modality Large Language Models (MLLMs) such as LLaVA\cite{liu2024visual} and InstructBLIP\cite{instructblip}, which have demonstrated vast potential across various fields.

Despite their impressive vision-language understanding capabilities, MLLMs are still not considered trustworthy AI systems\cite{li2023trustworthy}. Previous research has shown that these models may produce inconsistent responses to input images\cite{liu2023mitigating}, \cite{li2023comprehensive}, a phenomenon commonly referred to as `hallucinations'. One of the contributing factors to these hallucinations is the limited self-awareness of models. They lack a clear distinction between `what they know' and `what they do not know'. While this issue has been explored in high-level visual tasks \cite{wang2024mm}, the challenge of hallucinations in low-level visual tasks remains largely unexplored.

Meanwhile, the self-awareness of MLLMs in low-level visual perception and understanding plays significant roles in image quality assessment (IQA) \cite{hosu2020koniq, fang2020perceptual} and related tasks, including perceptual distortions (noise, blur) \cite{su2021koniq++, wu2023towards}, as well as other low-level attributes (color, lighting, composition, and style) \cite{kong2016photo}. These factors are relate to natural photo aesthetics \cite{murray2012ava} and AI-generated images \cite{li2023agiqa, xu2024imagereward}. These low-level visual abilities are strongly associated with a wide range of applications, such as recommendation \cite{wu2023exploring} or visual quality enhancement \cite{zhang2018unreasonable}. Therefore, it is crucial to evaluate the self-awareness capabilities of these general foundation models in low-level visual perception and understanding, and the response confidence of models to specific low-level tasks in order to reduce the reliance on extensive human resources.

In this paper, we propose the first systematic benchmark QL-Bench to measure the self-awareness ability of MLLMs in low-level visual tasks, which is constructed around a key question:

How is the self-awareness capability of MLLMs when simulating human low-level visual perception and understanding?

As illustrated in Fig. \ref{example}, similar to human behavior, the self-awareness of MLLMs should enable them to provide accurate answers when confident. In critical situations, when the question surpasses their comprehension or the available visual information, they should acknowledge their uncertainty rather than offering incorrect answers. Furthermore, we not only explore the self-awareness ability of MLLMs for three types of questions on a single image, but also extend low-level visual questions to image pairs. We hope that MLLMs can correctly answer low-level questions about images or respond with rejection when unsure how to answer.

In summary, we propose QL-Bench, a benchmark for the self-awareness of MLLMs on low-level visual tasks. Our contributions can be summarized as follows:

\begin{itemize}
\item We establish a baseline with 15 popular MLLMs for the self-awareness in low-level vision. To achieve this, we construct the LLSAVisionQA dataset, containing 2,990 single images and 1,999 image pairs with one low-level-related question-answer pair for each image. The question-answer pairs encompass three types of questions and focus on multiple low-level attributes to ensure diversity.
\item Comparison between single-image and multi-image tasks. Most models display stronger self-awareness in multi-image tasks compared to single-image tasks, although the accuracy of their responses decreased.
\item Comparison across question types. The models demonstrate weaker self-awareness in response to simple questions but exhibit greater self-awareness when dealing with more complex questions.
\end{itemize}

While these models demonstrate their ability to process information within their knowledge base, they tend to hesitate when recognizing the limitations of their understanding. These results suggest that MLLMs require clear strategies to support their ability to identify and acknowledge the boundaries of their knowledge.

\begin{figure}[!t]
\centerline{\includegraphics[width=3.5in]{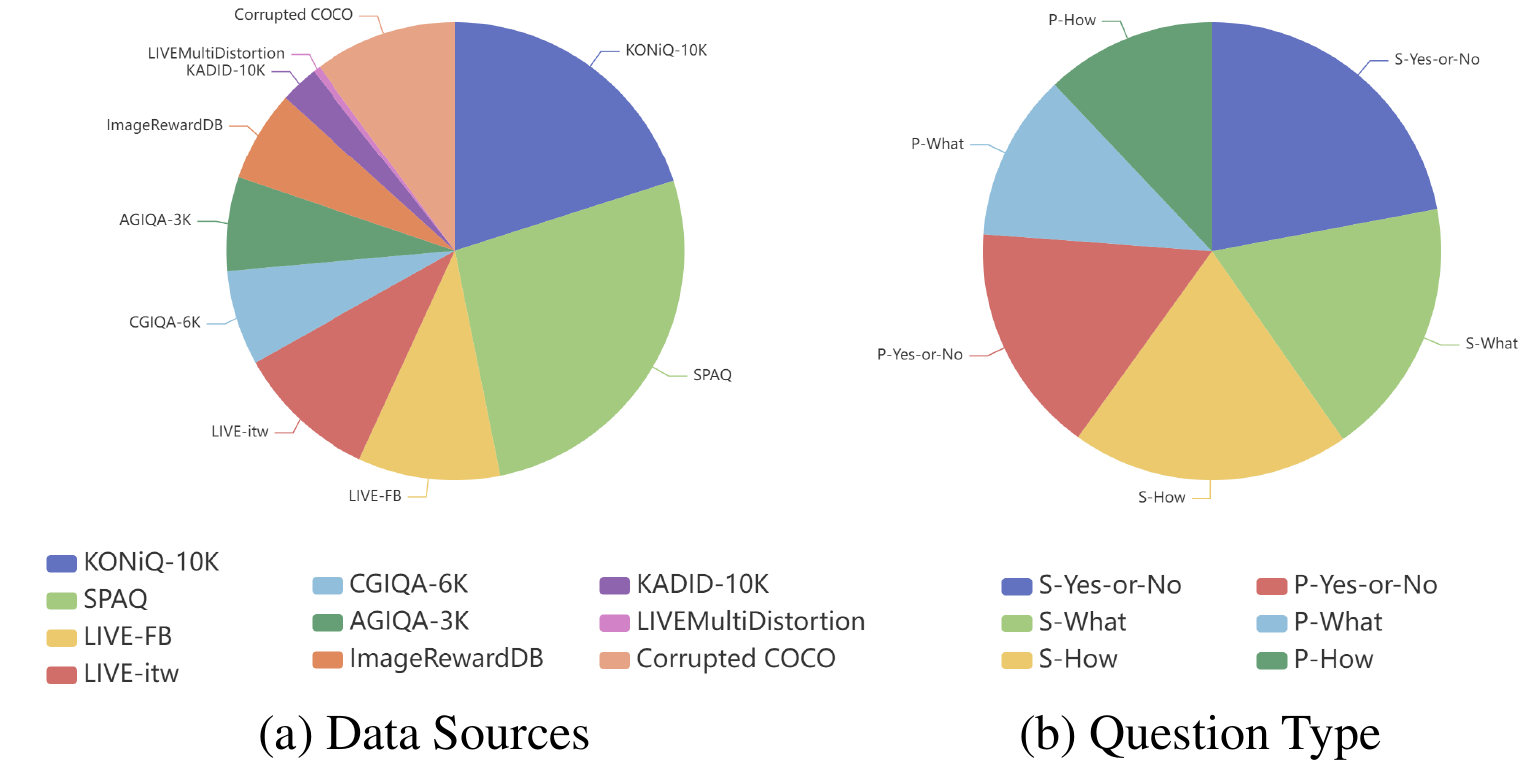}}
\caption{Overview of LLSAVisionQA dataset. (a) Data Sources: represents the sources of 2,990 individual images and 1,999 image pairs. (b) Question Type: shows the proportion of three types of questions.}
\label{dataset}
\vspace{-1em}
\end{figure}

\begin{table*}[!t]
    \centering
    \sisetup{table-format=2.3, detect-weight=true, mode=text, table-number-alignment=center}
    \caption{Results on the LLSAVisionQA for the low-level perception self-awareness ability of MLLMs. The best results are presented in \textbf{bold} and second/third \underline{underlined}.}
    \vspace{-0.5em}
    \label{qbench}
    \setstretch{1.1}
    \resizebox{0.85\textwidth}{!}{   
    \begin{tabular}{l S S S S S S S S S}
    \toprule[1.5pt]
    \multicolumn{1}{c}{\multirow{2}*{QL-Bench}} & \multicolumn{2}{c}{Yes-or-No} & \multicolumn{2}{c}{What} & \multicolumn{2}{c}{How} & \multicolumn{3}{c}{Total}\\ 
    \cmidrule(lr){2-3} \cmidrule(lr){4-5} \cmidrule(lr){6-7} \cmidrule(lr){8-10}
        & {$score_{cc}\uparrow$} & {$score_{rc}\uparrow$} & {$score_{cc}\uparrow$} & {$score_{rc}\uparrow$} & {$score_{cc}\uparrow$} & {$score_{rc}\uparrow$} & {$score_{cc}\uparrow$} & {$score_{rc}\uparrow$} & {$score_{sa}\uparrow$} \\
        \midrule

    LLaVA-Next (8B)\cite{liu2023improved}              & 65.392        &0.000            & 68.784      & 1.095      & 60.572     &0.000         & 64.849    & 0.334     & 65.183    \\
    LLaVA-Next (13B)\cite{liu2023improved} & 65.483        & 0.091         & 60.789      & 1.533      & 59.040      & 0.306      & 61.940     & 0.602     & 62.542    \\
    mPLUG-Owl2 (LLaMA2-7B)\cite{ye2024mplug}             & 65.938        &0.000            & 54.984      & \hspace{0.5em}\underline{4.600}        & 56.691     & 2.247      & 59.565    & 2.140      & 61.705    \\
    mPLUG-Owl (Bloomz-7B) \cite{ye2023mplug}
 & 53.734        & \hspace{0.5em}\underline{0.546}         & 42.607      & 3.395      & 39.224     & 2.554      & 45.585    & 2.074     & 47.659    \\
    InstructBLIP (Vicuna-7B) \cite{instructblip}           & 61.566        &0.000            & 46.440       & 1.095      & 45.455     & 2.962      & 51.672    & 1.304     & 52.976    \\
    InstructBLIP (Flan-T5-XL)\cite{instructblip}          & 62.204        & 0.273         & 54.217      & 2.738      & 52.809     & 1.124      & 56.689    & 1.304     & 57.993    \\
    InternLM-XComposer2 (7B) \cite{internlmxcomposer2}
           & 52.732        &0.000            & 56.736      & \textbf{10.515}     & 45.250      & \textbf{11.951}     & 51.505    & \textbf{ 7.124}     & 58.629    \\
    InternLM-XComposer2d5 (7B) \cite{internlmxcomposer2_5}         & \textbf{73.133}        & 0.091         & \underline{72.618}      & 3.176      & \underline{63.432}     & 0.409      & \textbf{69.799}    & 1.137     & \textbf{70.936}    \\
    InternLM-XComposer2 (4KHD-7B) \cite{internlmxcomposer2_4khd}     & 68.488        &0.000            & \textbf{74.808}      & 0.438      & \textbf{65.986}     &0.000         & \underline{69.599}    & 0.134     & \underline{69.733}    \\
    InfiMM (Zephyr-7B) \cite{InfiMM}                   & 59.927        &0.000            & 53.779      & 2.081      & 44.433     & 0.919      & 52.977    & 0.936     & 53.913    \\
    Fuyu (Persimmon-8B) \cite{bavishi2023multimodal}    & 59.199        & 0.091         & 42.388      & 1.533      & 41.369     & 1.430      & 48.227    & 0.970      & 49.197    \\
    Emu2-Chat (LLaMA-33B) \cite{sun2024generative}                        & 68.488        &0.000            & 61.665      & 4.053      & 44.127     & \hspace{0.5em}\underline{6.231}      & 58.428    & \hspace{0.5em}\underline{3.278}     & 61.706    \\ \hdashline
    GPT-4V (Close-Source) \cite{openai2023gpt4}                       & \underline{68.579}        & 0.091         & 70.427      & 1.314      & \underline{60.572}     & 1.532      & 66.522    & 0.936     & 67.458    \\
    GPT-4O (Close-Source) \cite{openai2023gpt4}                        & \underline{69.581}        & \hspace{0.5em}\underline{1.457}         & \underline{71.303}      & 3.176      & 59.755     & 2.656      & \underline{66.890}     & 2.375     & \underline{69.265}    \\
    Gemini-1.5-Pro (Close-Source) \cite{google2023geminipro}  & 58.106        & \textbf{ 4.736}         & 55.969      & \hspace{0.5em}\underline{6.900}        & 47.804     & \hspace{0.5em}\underline{5.107}      & 54.080     & \hspace{0.5em}\underline{5.518}     & 59.598  \\ \bottomrule[1.5pt]
    \end{tabular}} 
\end{table*}

\begin{table*}[!t]
    \centering
    \sisetup{table-format=2.3, detect-weight=true, mode=text, table-number-alignment=center} 
    \caption{Results on the LLSAVisionQA for the low-level perception-pair self-awareness ability of MLLMs. The best results are presented in \textbf{bold} and second/third \underline{underlined}.}
    \vspace{-0.5em}
    \label{qbench+}
    \setstretch{1.1}
    \resizebox{0.85\textwidth}{!}{
    \begin{tabular}{l S S S S S S S S S}
    \toprule[1.5pt]
    \multicolumn{1}{c}{\multirow{2}*{QL-Bench}} & \multicolumn{2}{c}{Yes-or-No} & \multicolumn{2}{c}{What} & \multicolumn{2}{c}{How} & \multicolumn{3}{c}{Total}\\ 
    \cmidrule(lr){2-3} \cmidrule(lr){4-5} \cmidrule(lr){6-7} \cmidrule(lr){8-10}
        & {$score_{cc}\uparrow$} & {$score_{rc}\uparrow$} & {$score_{cc}\uparrow$} & {$score_{rc}\uparrow$} & {$score_{cc}\uparrow$} & {$score_{rc}\uparrow$} & {$score_{cc}\uparrow$} & {$score_{rc}\uparrow$} & {$score_{sa}\uparrow$} \\
        \midrule
    InfiMM (Zephyr-7B) \cite{InfiMM}                    & \underline{54.254}        &0.000            & 40.716      & 2.555      & 41.597     & 0.666      & 46.473    & 0.950      & 47.423    \\
    mPLUG-Owl (Bloomz-7B) \cite{ye2023mplug} & 45.993        & 1.233         & 31.516      & \hspace{0.5em}\underline{3.748}      & 39.767     & 1.331      & 39.870     & 2.001     & 41.871    \\
    Emu2-Chat (LLaMA-33B) \cite{sun2024generative}                        & 49.199        &0.000            & 34.753      & \hspace{0.5em}\underline{4.770}       & 39.767     & \hspace{0.5em}\underline{6.156}      & 42.121    & \hspace{0.5em}\underline{3.252}     & 45.373    \\  \hdashline
    GPT-4V (Close-Source) \cite{openai2023gpt4}                            & \underline{59.556}        & \hspace{0.5em}\underline{2.096}         & \underline{59.966}      & 1.022      & \underline{58.403}     & 1.331      &\underline{59.330}    & 1.551     & \underline{60.881}    \\
    GPT-4O (Close-Source) \cite{openai2023gpt4}                            & \textbf{63.132}        & \hspace{0.5em}\underline{2.713}         & \textbf{62.521}      & 0.852      & \textbf{62.729}     & \hspace{0.5em}\underline{3.661}      & \textbf{62.831}    & \hspace{0.5em}\underline{2.451}     & \underline{65.282}    \\
    Gemini-1.5-Pro (Close-Source) \cite{google2023geminipro}                   & 52.035        & \textbf{14.797}        & \underline{45.826}      & \textbf{14.310}      & \underline{45.258}     & \textbf{35.108}     & \underline{48.174}    & \textbf{20.760}     & \textbf{68.934}  \\     \bottomrule[1.5pt]
    \end{tabular}
    }
    \vspace{-1em}
\end{table*}

\section{LLSAVisionQA Dataset}
To evaluate the self-awareness ability of different MLLMs under low-level visual features, we construct the LLSAVisionQA dataset, which includes 2,990 single images and 1,999 image pairs from 10 diverse sources, as shown in Fig. \ref{dataset}a, representing the image sources. Aligned with existing practices \cite{liu2023mmbench}, \cite{lu2023evaluation}, each single image or image pair in LLSAVisionQA is equipped with a question, alongside a correct answer, false answers and an `I don't know' option. Specifically, we design three diverse types of questions: `Yes-or-No' questions, `What' questions, and `How' questions, as illustrated in Fig. \ref{dataset}b, which shows the total number of each question type. 
The proposed LLSAVisionQA dataset focuses on distortion, color, overall, and other attributes through three types of questions, providing a comprehensive benchmark for the low-level perception self-awareness ability of MLLMs on single and paired images.

\section{Experiments}

\subsection{Evaluation Strategy}
Self-awareness involves the ability to recognize both ``knowns" and ``unknowns." To evaluate this capability in the QL-Bench, we introduce three metrics to measure the self-awareness of MLLMs.

\begin{itemize}
\item $score_{cc}$: It reflects the proportion of questions that the model answers correctly.
\item $score_{rc}$: It represents the proportion of questions that the model appropriately rejects.
\item $score_{sa}$: It is the sum of $score_{cc}$ and $score_{rc}$, representing the overall self-awareness of the model.
\end{itemize}

Before detailing the calculation of these metrics, we introduce some indicators to avoid confusion. For each question $q_i$ in the test set $q$, $c_i$ and $r_i$ represent the indices of the correct answer and the refusal option, respectively. Therefore, $score_{cc}$ and $score_{rc}$ can be defined as:
\begin{equation}
score_{cc} = \frac{100 \cdot \sum_{i=1}^{|q|} \mathbb{I}(p_i = c_i)}{|q|}
\end{equation}
\begin{equation}
score_{rc} = \frac{100 \cdot \sum_{i=1}^{|q|} \mathbb{I}(p_i = r_i) \cdot \mathbb{I}(q_i \text{ is unknown})}{|q|}
\end{equation}
where $p_i$ represents the prediction of the evaluated MLLMs for $q_i$. Since the model may refuse to answer questions that actually knows, when a refusal option is available, we address this issue by removing the refusal option `I don't know', forcing the model to select an answer. If the model chooses the correct answer, it demonstrates that the model actually knows the correct response. Consequently, $\mathbb{I}(q_i \text{ is unknown})$ can be defined as follows:

\begin{equation}
\mathbb{I}(q_i \text{ is unknown})=\mathbb{I}(p'_i \neq c_i \mid p_i = r_i) 
\end{equation}
where $p'_{i}$ represents prediction of the model without the refusal option. The self-awareness score($score_{sa}$) is calculated as:
\begin{equation}
score_{sa} = score_{cc} + score_{rc} 
\end{equation}

\begin{table*}[!t]
    \centering
    \sisetup{
      table-format=3.3,       
      detect-weight=true,     
      mode=text,
      detect-inline-weight=math, 
      table-number-alignment=center,
      table-space-text-post = \% 
    }
    \caption{Results of Answer Rate and Answer Accuracy of MLLMs on LLSAVisionQA for the low-level perception self-awareness ability. The best results are presented in \textbf{bold} and second/third \underline{underlined}.}
    \vspace{-0.5em}
    \label{qbenchACC}
    \setstretch{1.2}
    \resizebox{0.85\textwidth}{!}{
    \begin{tabular}{l S S S S S S}
    \toprule[1.5pt]
    \multicolumn{1}{c}{\multirow{2}{*}{QL-Bench}}   & \multicolumn{2}{c}{Yes-or-No} & \multicolumn{2}{c}{What} & \multicolumn{2}{c}{How}  \\
    \cmidrule(lr){2-3} \cmidrule(lr){4-5} \cmidrule(lr){6-7}
                                     & {Answer Acc$\uparrow$}    & {Answer Rate$\uparrow$}   & {Answer Acc$\uparrow$} & {Answer Rate$\uparrow$} & {Answer Acc$\uparrow$} & {Answer Rate$\uparrow$} \\ \midrule
    LLaVA-Next (8B)\cite{liu2023improved}               & 65.392\%       & \textbf{100.000\%}           & 69.546\%     & \hspace{0.5em}\underline{98.905\%}     & 60.572\%     & \textbf{100.000\%}         \\
    LLaVA-Next (13B)\cite{liu2023improved}              & 65.542\%        & \hspace{0.5em}\underline{99.909\%}        & 61.735\%     & 98.467\%      & 59.221\%     & \hspace{0.5em}\underline{99.694\%}      \\
    mPLUG-Owl2 (LLaMA2-7B)\cite{ye2024mplug}               & 65.938\%        & \textbf{100.000\%}           & 57.635\%     & 95.400\%        & 57.994\%     & 97.753\%      \\
    mPLUG-Owl (Bloomz-7B) \cite{ye2023mplug} & 54.029\%        & 99.454\%        & 44.104\%     & 96.605\%      & 40.252\%     & 97.446\%      \\
    InstructBLIP (Vicuna-7B)\cite{instructblip}           & 61.566\%        & \textbf{100.000\%}           & 46.955\%     & \hspace{0.5em}\underline{98.905\%}      & 46.842\%     & 97.038\%      \\
    InstructBLIP (Flan-T5-XL)\cite{instructblip}          & 62.374\%        & \hspace{0.5em}\underline{99.727\%}        & 55.743\%     & 97.262\%      & 53.409\%     & 98.876\%      \\
    InternLM-XComposer2 (7B) \cite{internlmxcomposer2}           & 52.732\%        & \textbf{100.000\%}           & 63.403\%     & 89.485\%      & 51.392\%     & 88.049\%      \\
    InternLM-XComposer2d5 (7B) \cite{internlmxcomposer2_5}         & \hspace{0.6em}\textbf{73.200\%}          & \hspace{0.5em}\underline{99.909\%}        & \hspace{0.5em}\underline{75.000\%}         & 96.824\%      & \hspace{0.5em}\underline{63.692\%}     & \hspace{0.5em}\underline{99.591\%}      \\
    InternLM-XComposer2 (4KHD-7B) \cite{internlmxcomposer2_4khd}      & 68.488\%        & \textbf{100.000\%}           & \hspace{0.6em}\textbf{75.138\%}     & \hspace{0.6em}\textbf{99.562\%}      & \hspace{0.6em}\textbf{65.986\%}     & \textbf{100.000\%}         \\
    InfiMM (Zephyr-7B) \cite{InfiMM}                    & 59.927\%       & \textbf{100.000\%}           & 54.922\%     & 97.919\%      & 44.845\%     & 99.081\%      \\
    Fuyu (Persimmon-8B) \cite{bavishi2023multimodal}                          & 59.253\%        & \hspace{0.5em}\underline{99.909\%}        & 43.048\%     & 98.467\%      & 41.969\%     & 98.570\%       \\
    Emu2-Chat (LLaMA-33B) \cite{sun2024generative}                        & 68.488\%        & \textbf{100.000\%}           & 64.269\%     & 95.947\%      & 47.059\%     & 93.769\%      \\  \hdashline
    GPT-4V (Close-Source) \cite{openai2023gpt4}                            & \hspace{0.5em}\underline{68.642\%}        & \hspace{0.5em}\underline{99.909\%}        & 71.365\%     & 98.686\%      & \hspace{0.5em}\underline{61.515\%}     & 98.468\%      \\
    GPT-4O (Close-Source) \cite{openai2023gpt4}                            & \hspace{0.5em}\underline{70.610\%}         & 98.543\%        & \hspace{0.5em}\underline{73.643\%}     & 96.824\%      & 61.385\%     & 97.344\%      \\
    Gemini-1.5-Pro (Close-Source) \cite{google2023geminipro}                   & 60.994\%        & 95.264\%        & 60.118\%     & 93.100\%        & 50.377\%     & 94.893\%    \\ \bottomrule[1.5pt] 
    \end{tabular}
    }
\end{table*}

\begin{table*}[!t]
    \centering
    \sisetup{
      table-format=3.3,       
      detect-weight=true,     
      mode=text,
      detect-inline-weight=math, 
      table-number-alignment=center,
      table-space-text-post = \%  
    }
    \caption{Results of Answer Rate and Answer Accuracy of MLLMs on LLSAVisionQA for the low-level perception-pair self-awareness ability. The best results are presented in \textbf{bold} and second/third \underline{underlined}.}
    \vspace{-0.5em}
    \label{qbench+ACC}
    \setstretch{1.2}
    \resizebox{0.85\textwidth}{!}{
    \begin{tabular}{l S S S S S S}
    \toprule[1.5pt]    
    \multicolumn{1}{c}{\multirow{2}{*}{QL-Bench}} & \multicolumn{2}{c}{Yes-or-No} & \multicolumn{2}{c}{What} & \multicolumn{2}{c}{How}  \\
    \cmidrule(lr){2-3} \cmidrule(lr){4-5} \cmidrule(lr){6-7}
                                     & {Answer Acc$\uparrow$}    & {Answer Rate$\uparrow$}   & {Answer Acc$\uparrow$} & {Answer Rate$\uparrow$} & {Answer Acc$\uparrow$} & {Answer Rate$\uparrow$} \\ \midrule
    InfiMM (Zephyr-7B) \cite{InfiMM}                                        & 54.254\%        & \textbf{100.000\%}           & 41.783\%     & \hspace{0.5em}\underline{97.445\%}      & 41.876\%     & \hspace{0.6em}\textbf{99.334\%}      \\
    mPLUG-Owl (Bloomz-7B) \cite{ye2023mplug}                     & 46.567\%        & \hspace{0.5em}\underline{98.767\%}        & 32.743\%     & 96.252\%      & 40.304\%     & \hspace{0.5em}\underline{98.669\%}      \\
    Emu2-Chat (LLaMA-33B) \cite{sun2024generative}                                            & 49.199\%        & \textbf{100.000\%}           & 36.494\%     & 95.230\%       & 42.376\%     & 93.844\%      \\  \hdashline
    GPT-4V (Close-Source) \cite{openai2023gpt4}                                                & \hspace{0.5em}\underline{60.831\%}        & \hspace{0.5em}\underline{97.904\%}        & \hspace{0.5em}\underline{60.585\%}     & \hspace{0.5em}\underline{98.978\%}      & \hspace{0.5em}\underline{59.191\%}     & \hspace{0.5em}\underline{98.669\%}      \\
    GPT-4O (Close-Source) \cite{openai2023gpt4}                                                & \hspace{0.6em}\textbf{64.892\%}        & 97.287\%        & \hspace{0.6em}\textbf{63.058\%}     & \hspace{0.6em}\textbf{99.148\%}      & \hspace{0.5em}\underline{65.112\%}     & 96.339\%      \\
    Gemini-1.5-Pro (Close-Source) \cite{google2023geminipro}                                       & \hspace{0.5em}\underline{61.071\%}        & 85.203\%        & \hspace{0.5em}\underline{53.479\%}     & 85.690\%       & \hspace{0.6em}\textbf{69.744\%}     & 64.892\%      \\ \bottomrule[1.5pt]
    \end{tabular}
    }
    \vspace{-1em}
\end{table*}

\subsection{Main Results}
We evaluate fifteen popular MLLMs on the QL-Bench, including twelve open-source models: LLaVA-Next (8B)\cite{liu2023improved}, LLaVA-Next (13B)\cite{liu2023improved}, mPLUG-Owl2\cite{ye2024mplug}, mPLUG-Owl\cite{ye2023mplug}, InstructBLIP (Vicuna-7B) \cite{instructblip}, InstructBLIP (Flan-T5-XL) \cite{instructblip}, InternLM-XComposer2 \cite{internlmxcomposer2}, InternLM-XComposer2d5 \cite{internlmxcomposer2_5}, InternLM-XComposer2 (4KHD) \cite{internlmxcomposer2_4khd}, InfiMM \cite{InfiMM}, Fuyu\cite{bavishi2023multimodal}, Emu2-Chat \cite{sun2024generative} and three close-source models: GPT-4V \cite{openai2023gpt4}, GPT-4O \cite{openai2023gpt4}, and Gemini-1.5-Pro \cite{google2023geminipro}. The self-awareness scores of these MLLMs on single-image tasks are shown in Table \ref{qbench}. Overall, InternLM-XComposer2d5\cite{internlmxcomposer2_5} achieves the best $score_{sa}$. Gemini-1.5-Pro\cite{google2023geminipro} shows the best $score_{rc}$ for `Yes-or-No' questions, while InternLM-XComposer2\cite{internlmxcomposer2} demonstrates the best $score_{rc}$ for `What' and `How' questions. For $score_{cc}$, InternLM-XComposer2d5\cite{internlmxcomposer2_5} achieves the best in `Yes-or-No' questions, and InternLM-XComposer2 (4KHD) \cite{internlmxcomposer2_4khd} shows the best in `What' and `How' questions. We observe that most models are not able to balance both $score_{cc}$ and $score_{rc}$ metrics effectively, and their ability to detect the boundaries of low-level visual knowledge is very limited, especially in simple `Yes-or-No' questions.

At the same time, in order to explore the self-awareness capability of MLLMs on multi-image tasks, we evaluate six widely used MLLMs, including three open-source models: InfiMM \cite{InfiMM}, mPLUG-Owl \cite{ye2023mplug}, Emu2-Chat \cite{sun2024generative} and three close-source models: GPT-4V \cite{openai2023gpt4}, GPT-4O \cite{openai2023gpt4}, and Gemini-1.5-Pro \cite{google2023geminipro}. The self-awareness scores of these MLLMs on multi-image tasks are shown in Table \ref{qbench+}. On the whole, close-source MLLMs are more robust in this task. GPT-4O\cite{openai2023gpt4} and GPT-4V\cite{openai2023gpt4} outperform other models significantly in terms of $score_{cc}$, showing a 5\% and 9\% improvement over the third-highest model for `Yes-or-No' questions, a 14\% and 17\% improvement for `What' questions, and an 13\% and 17\% improvement for `How' questions. On the other hand, Gemini-1.5-Pro \cite{google2023geminipro} shows far superior $score_{rc}$ compared to other models in all three categories of questions, with a 12\% improvement over the second-highest model for `Yes-or-No' questions, a 10\% improvement for `What' questions, and a 29\% improvement for `How' questions. Similar to single-image tasks, most MLLMs struggle to balance $score_{cc}$ and $score_{rc}$ when dealing with multi-image tasks. However, for the same type of questions, MLLMs exhibit an improved ability to perceive the boundaries of low-level visual knowledge when handling multi-image tasks.

\subsection{Refusal Behavior of MLLMs}
In order to provide a more comprehensive analysis, we introduce two additional metrics to represent the refusal behavior of MLLMs.
\begin{equation}
\text{Answer Acc} = \frac{\sum_{i=1}^{|q|} \mathbb{I}(p_i = c_i)}{\sum_{i=1}^{|q|} \mathbb{I}(p_i \neq r_i)}
\end{equation}
\begin{equation}
\text{Answer Rate} = \frac{\sum_{i=1}^{|q|} \mathbb{I}(p_i \neq r_i)}{|q|}
\end{equation}
where Answer Accuracy represents the proportion of correct predictions among the questions, which model chooses to answer. Answer Rate reflects the proportion of all questions that the model attempts to answer.

Tables \ref{qbenchACC} and \ref{qbench+ACC} present the answer accuracy and answer rate of MLLMs in single-image tasks and image-pair tasks, respectively. The results show that most MLLMs achieve the answer rate of 100\% or close to 100\% for `Yes-or-No' questions. InternLM-XComposer2 (7B)\cite{internlmxcomposer2} and Gemini-1.5-Pro\cite{google2023geminipro} demonstrate relatively low answer rate in single-image and multi-image tasks, indicating the strongest capability to recognize unknown content and the highest level of self-awareness. Overall, when examining the answer accuracy, it is evident that current models face challenges in accurately identifying unknown low-level visual information, highlighting considerable potential for further improvements.

\section{Conclusion}
In this research, we introduce QL-Bench, a benchmark designed to evaluate the self-awareness of MLLMs in low-level visual tasks. We evaluate MLLMs from three types of questions: `Yes-or-no', `What', and `How'. In addition, recognizing the importance of identifying differences and similarities in image pairs, our benchmark includes both single-image and image-pair tasks focused on low-level visual perception and understanding. Our results indicate that, despite not receiving specialized training, some models show strong low-level visual capabilities when processing images. However, their self-awareness remains relatively underdeveloped. The same model tends to answer simpler questions more accurately than complex ones, although its self-awareness is weaker in these simpler questions. We hope that the insights gained from QL-Bench will contribute to the further advancement of MLLMs, particularly in enhancing their self-awareness when it comes to perceiving and understanding low-level visual elements.
\bibliographystyle{IEEEtran}
\bibliography{ref}

\end{document}